%% file: adviz.tex
\documentclass{article}
\usepackage{ijcai18}
\usepackage{natbib}
\usepackage{times}
\usepackage{xcolor}
\usepackage{soul}
\usepackage[utf8]{inputenc}
\usepackage[small]{caption}
\usepackage{subcaption}

\usepackage{algorithm}
\usepackage{algpseudocode}
\usepackage{amsmath,amssymb,amsfonts}
\usepackage{graphicx}
\usepackage{xcolor}
\usepackage{todonotes}
\usepackage{booktabs}
\usepackage{textcomp}
\usepackage{xspace}
\usepackage{multirow}
\usepackage{comment}
\usepackage{paralist}
\usepackage{multirow}
\usepackage{wasysym}

\usepackage{url}
\def\BibTeX{{\rm B\kern-.05em{\sc i\kern-.025em b}\kern-.08em
		T\kern-.1667em\lower.7ex\hbox{E}\kern-.125emX}}

\newcommand{\vx}{\mathbf{x}}
\newcommand{\vtheta}{\mathbf{\theta}}

\newcommand{\expect}{\mathbb{E}}
\newcommand{\distrib}{\mathcal{D}}
\newcommand{\binaryset}{\{0,1\}}

\newcommand{\fgsm}{\texttt{FGSM}}
\newcommand{\mfgsm}{$\fgsm^k$}
\newcommand{\rmfgsm}{\texttt{r}\mfgsm}
\newcommand{\dmfgsm}{\texttt{d}\mfgsm}

\newcommand{\mBCA}{$\texttt{BCA}^k$}
\newcommand{\mBGA}{$\texttt{BGA}^k$}

\title{
	On Visual Hallmarks of Robustness to Adversarial Malware}

\author{
Alex Huang, 
Abdullah Al-Dujaili, 
Erik Hemberg, 
Una-May O'Reilly
\\ 
CSAIL, MIT, USA \\
alhuang@mit.edu,
aldujail@mit.edu,
hembergerik@csail.mit.edu,
unamay@csail.mit.edu
}

\begin{document}

\maketitle
\input{abstract}
\input{introduction}
\input{background}

\input{tools}
\input{conclusions}

\newpage

\bibliographystyle{named}
\bibliography{ijcai18}

\end{document}

%% file: abstract.tex
\begin{abstract}
A central challenge of adversarial learning is to interpret the resulting hardened model. In this contribution, we ask how robust generalization can be visually discerned and whether a concise view of the interactions between a hardened decision map and input samples is possible.  We first provide a means of visually comparing a hardened model's loss behavior with respect to the adversarial variants generated during training versus loss behavior with respect to adversarial variants generated from other sources.  This allows us to confirm that the association of observed flatness of a loss landscape with generalization that is seen with naturally trained models extends to adversarially hardened models and robust generalization.  To complement these means of interpreting model parameter robustness we also use self-organizing maps to provide a visual means of superimposing adversarial and natural variants on a model's decision space, thus allowing the model's global robustness to be comprehensively examined. 
\end{abstract}

%% file: introduction.tex
\section{Introduction}
\label{sec:intro}

Neural Network~(NN) models are vulnerable to \textit{adversarial variants} \citep{adversarial_paper, another_adversarial_paper, rl_adversarial_paper}. An adversarial variant is a data sample generated by making a small modification to an existing sample such that the natural (original) and new adversarial variant are very similar (e.g. indistinguishable to the human eye for images) but the adversarial variant is incorrectly classified. This was first studied in the context of image classification \citep{adversarial_paper}, but has also been studied in malware detection~\citep{adversarial_malware}.

We are interested in how to interpret and visually assess adversarially hardened NN models.  One existing visualization is \textit{loss progression}, see Figure~\ref{loss_progression}~\citep{madry2017towards}. Assuming a saddle-point formulation, it plots the loss of a hardened or naturally trained model given inputs that are adversarially generated variations of a single sample over iterations of inner maximization. Related to loss progression is \textit{loss evolution}, for example \citep[Fig. 5]{madry2017towards}. It plots the decrease in average inner maximization loss for the maximally adversarial variation of every sample per training step.  \textit{Loss histograms} are another interpretive tool, see Figure~\ref{histograms}~\citep{madry2017towards}. They differentiate a model's loss to adversarial variants for which it was retrained from loss to other sample sources in terms of the frequency of sample losses. Finally, \textit{decision boundary analysis} \citep{decision_boundary_analysis} can show how hardening a model affects the average distance from samples to the closest decision boundary (i.e. the point where the label of a sample changes). Distance is calculated in terms of iterations by perturbing a sample by a fixed amount each iteration until its label changes.

In contrast to the aforementioned tools, \textit{loss landscape} visualization reveals the geometry of the loss landscape around a model's parameters.  For models trained with natural samples, flatness (sharpness) has been associated with good (poor) generalization~\citep{landscape_shape_one,landscape_shape_two}. This \emph{natural}  generalization is subsumed by \emph{robust} generalization. Robust generalization is required for a hardened NN model because the model must handle benign and malicious samples plus  adversarial variations.  In this contribution we use an existing loss landscape tool by~\citet{filterwise} to visualize a loss landscape of a hardened model given the adversarial variations with which it was trained. Our aim is to then compare and contrast loss landscapes of models hardened by different adversarial learning methods. Furthermore, for each model hardened by a specific adversarial learning method, we aim to compare its loss landscape given adversarial variations derived during learning to its loss landscape given a bigger set of adversarial variations, including ones derived from adversarially learning another model with a different method.  This allows us to answer the general question of whether the flatness/sharpness association holds for robust generalization.\footnote{Standard generalization refers to the model's performance on previously unseen data. In the context of robust generalization above, we consider the model's performance on previously unseen data and unseen attacks.}

In considering input space interpretation, rather than parameter interpretation, \emph{blind spot coverage}, a scalar value computed during training, has been reported for NN models that have been trained with natural and adversarially generated samples in the binary input space~\citep{al2018adversarial}. A blind spot is, informally, a region in the input space where there is a lack of training examples. While blind spot coverage is related to the size of the input space region that has been extrapolated by the model's  decision map, it does not provide decision map-sample locality information. We aim for a way of indicating \textit{where} the adversarial variations  or benign and malicious samples are located relative to the decision boundary.

Towards this combined set of aims, this paper presents an adversarial loss landscape method and a decision, self-organizing map method that help interpret a hardened NN model.  We demonstrate the methods on the adversarial malware detection problem for portable executable (PE) files. We use models adversarially trained by 4 different methods and we assess each model given benign and malicious samples as well as adversarial variations arising either from its own hardening or from the hardening of the other 3 models.  In Section~\ref{sec:background} we formalize the adversarial malware detector hardening problem and describe the dataset we use. Section~\ref{sec:tools} describes the visualization tools. Finally, Section~\ref{sec:conclusion} outlines the conclusions and future work.

%% file: background.tex
\section{Formal Background}
\label{sec:background}

In this section, we briefly describe the problem of hardening machine learning malware detectors (binary classifiers) via adversarial learning and the setup used to train them. We adopt the notation and setup used in~\citep{al2018adversarial}.

\paragraph{Binary Executable Representation.} Based on extracted static features, each binary executable is represented by a feature indicator vector $\vx=[x_1,\ldots,x_m] \in \mathcal{X}$. That is, $\mathcal{X}=\binaryset^{m}$ and $x_j$ is a binary value that indicates whether the $j$th feature is present or not. Labels are denoted by $y \in \mathcal{Y}=\binaryset$, where $0$ and $1$ denote benign and malignant (malicious) executables, respectively.

\begin{figure}[h]
	\centering
	\includegraphics[width=0.3\textwidth,trim={0 10cm 14cm 0cm},clip]{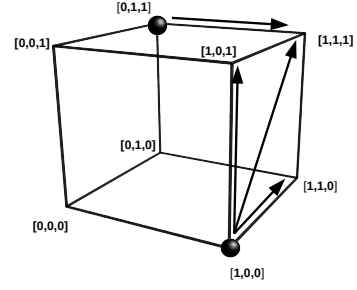} 
	\caption{The 3-dimensional feature indicator vector space of two malicious binary executables (malwares). The set of adversarial variations for the malware at $[1,0,0]$ is   $\mathcal{S}([1,0,0])=\{[1,0,0],[1,1,0],[1,0,1],[1,1,1]\}$, and for the malware at $[0,1,1]$ is $\mathcal{S}([0,1,1])=\{[0,1,1],[1,1,1]\}$. The arrows point to the set of allowed perturbations. Adapted from~\citep{al2018adversarial}.}
	\label{fig:syn-example}
\end{figure}

\paragraph{Adversarial Learning.} An adversarial malware variation  $\vx_{adv}$ (which may or may not be misclassified) of a correctly classified malware $\vx$ can be generated by perturbing $\vx$ in a way that preserves its malicious functionality and maximizes the loss $L$ of the binary classifier model of parameters $\theta\in \mathbb{R}^p$, i.e.,
\begin{equation}
\vx_{adv} \in  \arg\max_{\bar{\vx} \in \mathcal{S}(\vx)} L(\vtheta, \bar{\vx}, y=1)\;,
\label{eq:adv-samples}
\end{equation}
where $\mathcal{S}(\vx) \subseteq \mathcal{X}$ is the set of feature indicator vectors that preserve the functionality of malware $\vx$ (see Fig.~\ref{fig:syn-example}). 

In adversarial learning, adversarial variations are incorporated into the learning process in a saddle-point formulation. That is, we would like to find the optimal model parameters $\theta^*$ such that
\begin{equation}
\vtheta^* \in \underbrace{\arg\min_{\vtheta \in \mathbb{R}^p}\expect_{(\vx,y)\sim\distrib} \bigg[ \overbrace{\max_{\bar{\vx} \in \mathcal{S}(\vx)} \underbrace{L(\vtheta, \bar{\vx}, y)}_\text{natural loss}}^\text{adversarial loss}\bigg]}_\text{adversarial learning}\;.
\label{eq:saddle-problem}
\end{equation}

\begin{figure}[tb]
	\centering
	\includegraphics[width=0.4\textwidth]{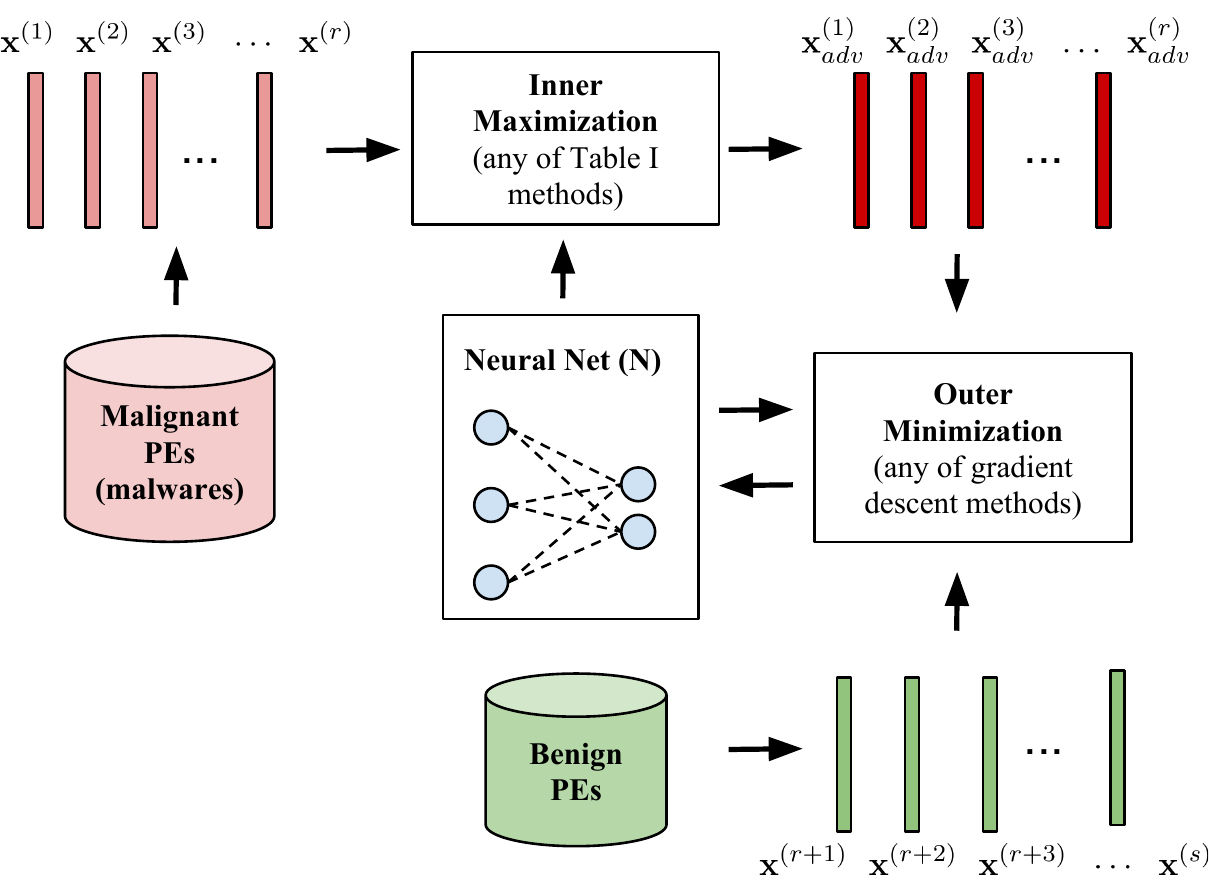}
	\caption{Overview of \textsc{Sleipnir} adversarial learning framework. Malware is perturbed by an inner maximization method to create adversarial variations. The generated adversarial variations and benign samples are used in an outer minimization problem. A neural network model is trained in batches by optimizing the adversarial and natural loss using gradient descent. Adapted from~\citep{al2018adversarial}.}
	\label{fig:RAMOverview}
\end{figure}

Solving the problem in Eq.~\eqref{eq:saddle-problem} involves an inner non-concave maximization problem and an outer non-convex minimization problem.
\cite{al2018adversarial} proposed a set of inner maximizer  algorithms---namely \rmfgsm, \dmfgsm, \mBGA, and \mBCA---and incorporated them in their \textsc{Sleipnir} framework~(see Fig.~\ref{fig:RAMOverview}) to solve Eq.~\eqref{eq:saddle-problem}.  In this context, the inner maximizers algorithms can also be regarded as adversarial (variant $\vx_{adv}$) generation methods. The framework was validated on a corpus of Windows portable executables as described next.

\paragraph{Dataset.}

The dataset is composed of Portable Executable (PE) files, a file format for executables in $32$-bit and $64$-bit Windows operating systems. The PE format encapsulates information necessary for Windows OS to manage the wrapped code, and this information can be extracted in order to construct feature representations. PE files are a natural choice due to their structure and widespread use as malware. The dataset used for visualization consists of $34,995$ malicious PE's from {{VirusShare}}\footnote{\url{https://virusshare.com/}} and $19,696$ benign PE's from {{CNET}}\footnote{\url{https://www.cnet.com/}}.  As described earlier, each PE is represented as a feature indicator vector, where each entry corresponds to a unique Windows API call, The value "1" in an entry denotes the presence of the corresponding API call. In our dataset, we observe a total of $22,761$ unique API calls, thus each PE file is represented as a  binary vector $\vx \in \mathcal{X}=\{0,1\}^{22761}$. We use the {LIEF}\footnote{\url{https://lief.quarkslab.com/}} library to parse each PE. Note that other established parsing tools can be used (e.g., \citep{pefile}) and we leave investigating different parsers for future work. The models were trained as described in~\citep{al2018adversarial}.

%% file: tools.tex
\section{Visualizing Adversarially Hardened Models}
\label{sec:tools}

In this section, we describe visualization tools that help elucidate blind spots and the robustness of hardened models.  We demonstrate these tools on  the four adversarially hardened models described in~\citep{al2018adversarial} in addition to the naturally trained model. The models are denoted by their inner maximizer (adversarial generation) methods: \dmfgsm, \rmfgsm , \mBGA,  \mBCA, and \texttt{Natural}, respectively. Based on~\citep{al2018adversarial}'s test set results, the order of the models from the most to the least robust is as follows: \rmfgsm, \mBGA, \dmfgsm, \mBCA, and \texttt{Natural}. Given this order, we ask the following question: would visualizing the loss landscape and the decision space based on the training set tell us something about the model's performance on the test set (i.e., its robust generalization)? We first present the loss progression and histograms of inner maxima employed by~\cite{madry2017towards}.

\subsection{Loss Progression of Inner Maxima Methods}

Loss progression plots can be used to show how well an adversarial generation method solves the inner maximization of~\eqref{eq:saddle-problem}. We reiterate that by \emph{inner maxima} (local or global), we mean the adversarial malware variations~$\vx_{adv}\in \mathcal{S}(\vx)$ generated when solving the inner maximization problem of Eq.~\eqref{eq:saddle-problem}. To generate a loss progression plot, we take a malware sample $\vx$ and track the model loss over iterations of the considered inner maximizer method. A low final loss value of an inner maximizer method indicates a failure to fool the model into thinking an adversarial variant is benign. 

\begin{figure*}[!htb]
	\centering
	\begin{subfigure}[t]{1.7in}
		\includegraphics[width=1.7in]{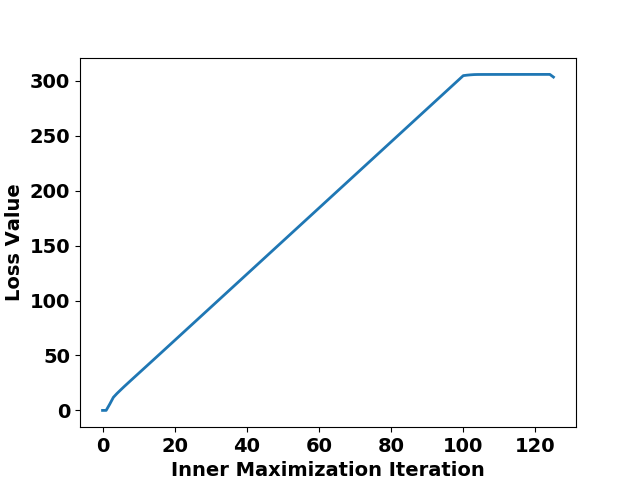}
		\caption{Model: \texttt{Natural} \\ Inner Maximizer: \dmfgsm}
		\label{natural_dfgsm}
	\end{subfigure}
	\begin{subfigure}[t]{1.7in}
		\includegraphics[width=1.7in]{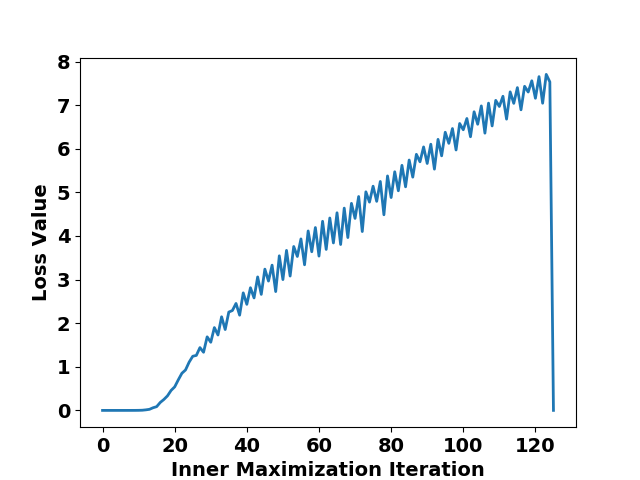}
		\caption{Model: \dmfgsm \\ Inner Maximizer: \dmfgsm}
		\label{dfgsm_dfgsm}
	\end{subfigure}
	\begin{subfigure}[t]{1.7in}
		\includegraphics[width=1.7in]{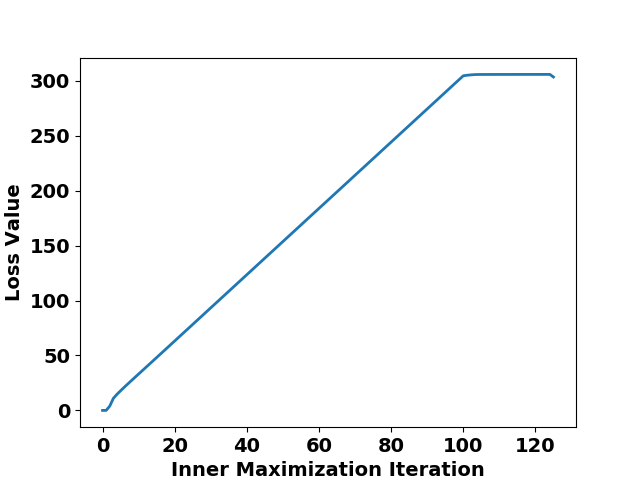}
		\caption{Model: \texttt{Natural} \\ Inner Maximizer: \rmfgsm}
		\label{natural_rfgsm}
	\end{subfigure}
	\begin{subfigure}[t]{1.7in}
		\includegraphics[width=1.7in]{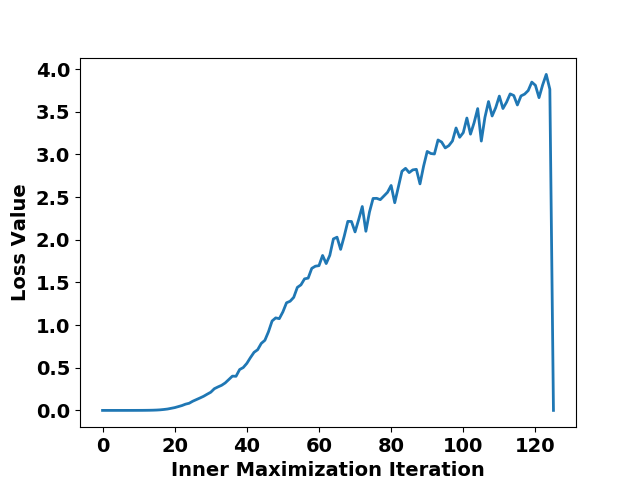}
		\caption{Model: \rmfgsm \\ Inner Maximizer: \rmfgsm}
		\label{rfgsm_rfgsm}
	\end{subfigure} \\
	\begin{subfigure}[t]{1.7in}
		\includegraphics[width=1.7in]{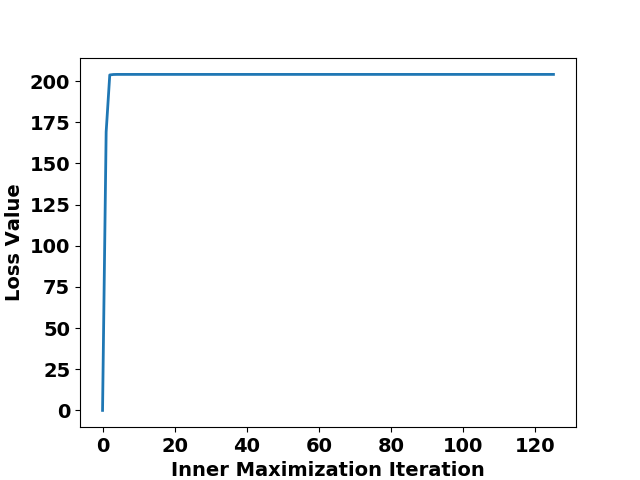}
		\caption{Model: \texttt{Natural} \\ Inner Maximizer: \mBGA }
		\label{natural_bga}
	\end{subfigure}
	\begin{subfigure}[t]{1.7in}
		\includegraphics[width=1.7in]{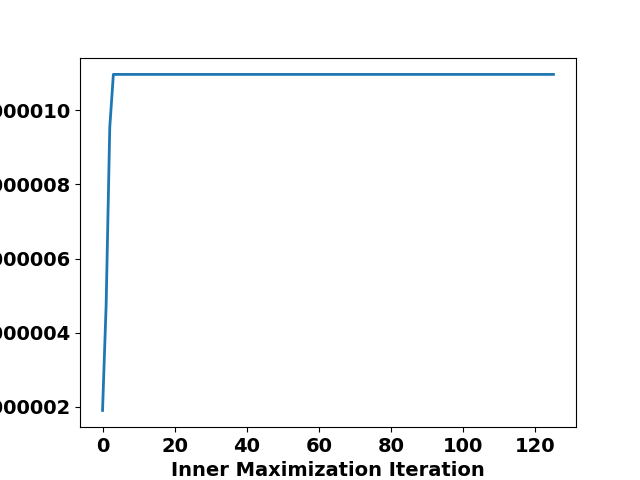}
		\caption{Model: \mBGA \\ Inner Maximizer: \mBGA }
		\label{bga_bga}
	\end{subfigure}
	\begin{subfigure}[t]{1.7in}
		\includegraphics[width=1.7in]{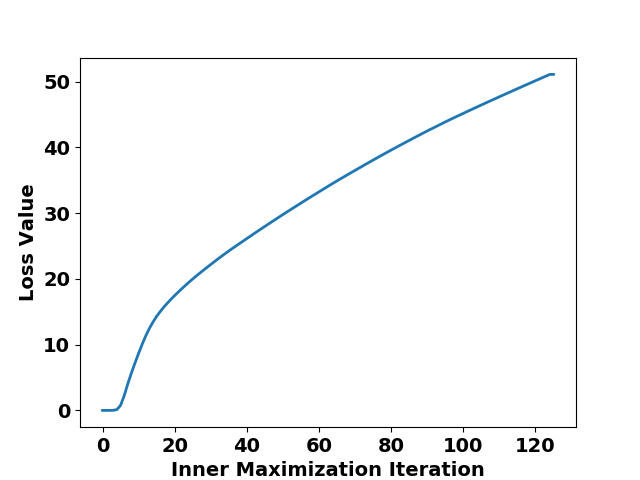}
		\caption{Model: \texttt{Natural} \\ Inner Maximizer: \mBCA}
		\label{natural_bca}
	\end{subfigure}
	\begin{subfigure}[t]{1.7in}
		\includegraphics[width=1.7in]{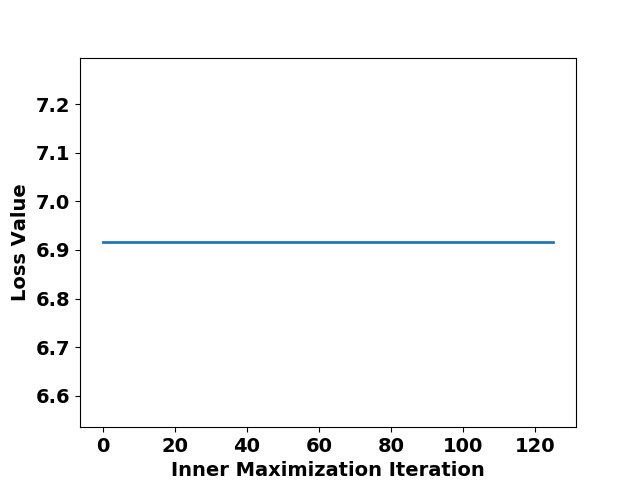}
		\caption{Model: \mBCA \\ Inner Maximizer: \mBCA}
		\label{bca_bca}
	\end{subfigure}
	\caption{Progressions of loss for a naturally trained model and a model trained with an adversarial generation method. Presentation of adversarial variation derfived from the generation method. The x~axis shows the number of iterations (steps) of the inner maximization.}
	\label{loss_progression}
\end{figure*}

In the plots in Figure~\ref{loss_progression}, the loss of the progressive variations of a malicious sample is shown. Each of the inner maximizer methods is able to inflict a high adversarial  loss on the naturally trained model (Figures \ref{natural_dfgsm}, \ref{natural_rfgsm}, \ref{natural_bga}, \ref{natural_bca}) but struggles to do so (i.e. inflicts a lower loss) on its adversarially hardened model (Figures \ref{dfgsm_dfgsm}, \ref{rfgsm_rfgsm}, \ref{bga_bga}, \ref{bca_bca}). Note the significantly smaller loss (y axis) scale of the adversarially hardened models in comparison to naturally trained models. 

Figures \ref{natural_rfgsm} and \ref{rfgsm_rfgsm} demonstrate the difference in loss between a naturally-trained model tested afterward againsts \rmfgsm~adversarial variations and an adversarially hardened model trained with \rmfgsm.
Against the \rmfgsm-trained model, \rmfgsm~variations increase the loss, but not enough to surmount the method's rounding threshold. This resets  the changes back to 0 at the last training iteration. We round the adversarial variations because our model accepts only binary inputs. 

\subsection{Loss Histograms of Inner Maxima Methods}

\begin{figure*}[!htb]
	\centering
	\begin{subfigure}[t]{1.35in}
		\includegraphics[width=1.35in]{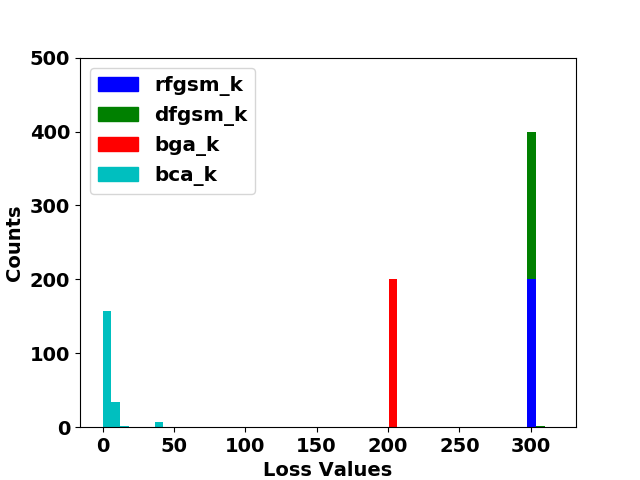}
		\caption{Naturally-Trained}
		\label{hist_natural}
	\end{subfigure}
	\begin{subfigure}[t]{1.35in}
		\includegraphics[width=1.35in]{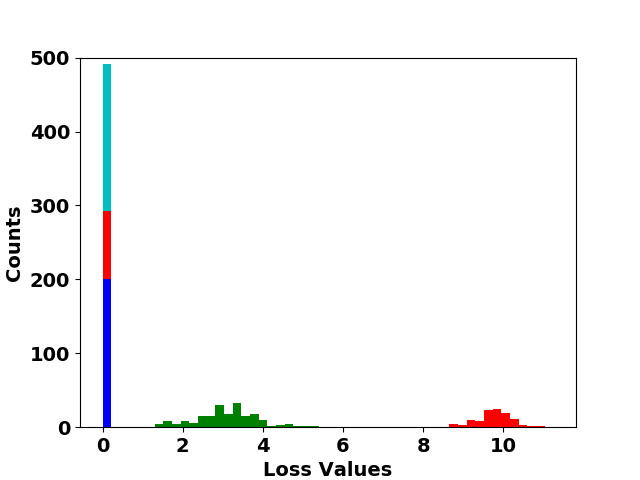}
		\caption{\dmfgsm-Trained}
		\label{hist_dfgsm}
	\end{subfigure}
	\begin{subfigure}[t]{1.35in}
		\includegraphics[width=1.35in]{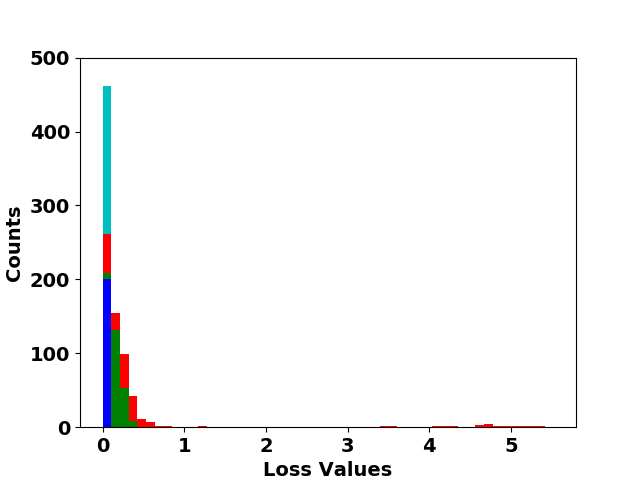}
		\caption{\rmfgsm-Trained}
		\label{hist_rfgsm}
	\end{subfigure}
	\begin{subfigure}[t]{1.35in}
		\includegraphics[width=1.35in]{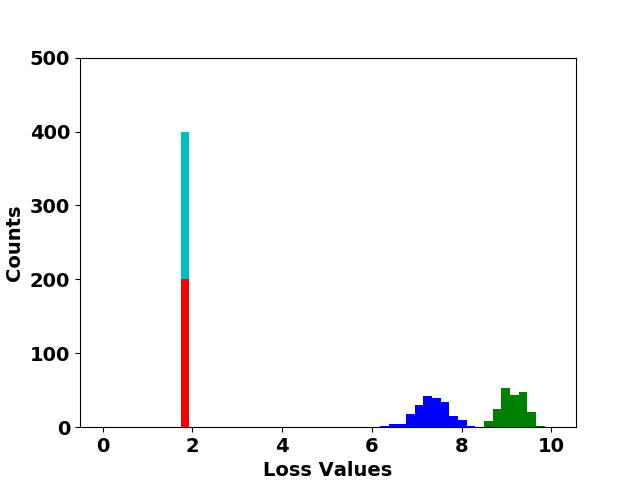}
		\caption{\mBGA-Trained}
		\label{hist_bga}
	\end{subfigure}
	\begin{subfigure}[t]{1.35in}
		\includegraphics[width=1.35in]{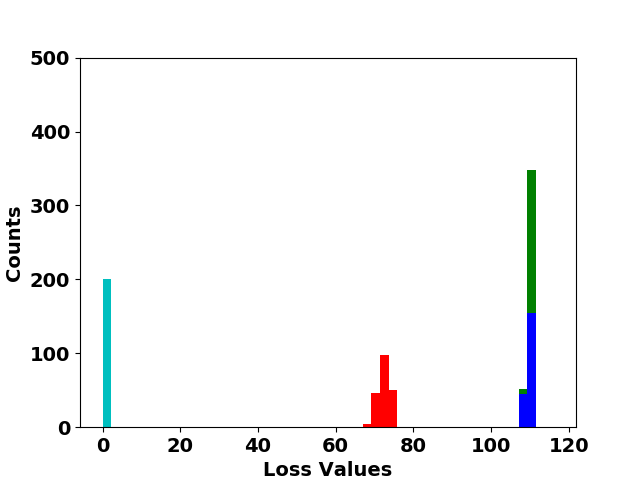}
		\caption{\mBCA-Trained}
		\label{hist_bca}
	\end{subfigure}
	\caption{Histograms of loss values from each type of adversarial generation method, blue is \rmfgsm, green is \dmfgsm, red is \mBGA, and light blue is \mBCA. The difference in x-axis scale among naturally and adversarially trained models shows that training with inner maximizer methods improves resistance to adversarial variants, with \dmfgsm, \rmfgsm, and \mBGA~more resistant than \mBGA.}	\label{histograms}
\end{figure*}

In addition to loss progression, the final loss values corresponding to different starting points $\vx^\prime \in \mathcal{S}(\vx)$ can be aggregated in a histogram. Figure \ref{histograms} illustrates the loss values for each model resulting from applying each of our adversarial generation methods (\dmfgsm, \rmfgsm, \mBGA, \mBCA) on a single malware $\vx$ sample along with 200 additional randomly sampled points in $\mathcal{S}(\vx)$. \dmfgsm, \rmfgsm, and \mBGA~demonstrate strong resistance, with loss values very close to 0 for all adversarial variations. \mBCA~on the other hand provides resistance only against itself, not any of the three other inner maximizers. This indicates that not all inner maximizers provide effective robustness against other inner maximizers, leading us to explore different techniques for visualizing this phenomenon. 

The loss progression and histogram plots visualize the model in a local way: solely in the context of a particular data sample. We would like to examine  the model in a global sense. To this end, we visualize the model's loss landscape and decision map.

\subsection{Loss Landscape}
To compare natural and hardened models in parameter space $\mathbb{R}^p$, we visualize the loss landscape. The loss landscape refers to the relationship between model parameters and loss values. This relationship is difficult to visualize due to the extreme high dimensionality of a NN model.  What can be plotted, however, is the immediate area around a particular set of parameters. The sharpness/flatness of this area is of great interest due to a hypothesized correlation between sharpness of the loss landscape and high generalization error \citep{landscape_shape_one,landscape_shape_two}. The goal of visualizing the loss landscape is to compare the effects of adversarial training. Toward this end, we plot the loss landscape in two ways: 1. using malicious and benign samples plus only adversarial variants generated with the same method used to train the model (Figure \ref{landscape_one_evasion}), and 2. using malicious and benign samples plus adversarial variants from all methods (Figure \ref{landscape_all_evasion}).

\paragraph{Method.} As described in Algorithm~\ref{loss_landscape_code}, line 3, we plot the loss landscape of our natural and hardened models using filter-wise normalization \citep{filterwise}. We use an range of -2 to 2 for $\alpha$ and $\beta$ with a 0.25 increment. Gaussian direction matrices $\delta$ and $\eta$ are generated separately for each plot. We use 250 samples for each loss calculation: benign and malicious samples, \dmfgsm-generated, \rmfgsm-generated, \mBGA-generated, and \mBCA-generated variants. Additionally, we use only benign and malicious samples from the model's training set for the loss calculation along with the adversarial variants.

\begin{algorithm}
	\tiny
	\caption{ \tiny Adversarial Loss Landscape \\
	\textbf{Requires:} \\
	$~$\hspace{\algorithmicindent}$\theta\in \mathbb{R}^{p}$: model parameters with $\theta^{(j)}\in \mathbb{R}^{p_j}$ as \\
	$~$\hspace{\algorithmicindent}$~$\hspace{\algorithmicindent} the  $j^{th}$ layer's parameters\\
	$~$\hspace{\algorithmicindent}$D^{bon}$: benign PEs dataset \\
	$~$\hspace{\algorithmicindent}$D^{mal}$: malicious PEs dataset \\
	$~$\hspace{\algorithmicindent}$D^{adv}$: adversarial malicious PEs dataset \\
	$~$\hspace{\algorithmicindent}$(\alpha_{min}, \alpha_{max})$: range of parameter $\alpha\in \mathbb{R}$\\
		$~$\hspace{\algorithmicindent}$(\beta_{min}, \beta_{max})$: range of parameter $\beta\in \mathbb{R}$
	}
	\label{loss_landscape_code}
	\begin{algorithmic}[1]
		\Statex {// filter-wise normalization~\citep{filterwise}}
		\For{each layer $j$ of the model's layers}
			\State $\delta^{(j)}_i, \eta^{(j)}_i \sim \mathcal{N}(0, 1)\;, \forall i \in [p_j]$
			\State Scale $\delta_i$ and $\eta_i$ to have the same $\ell_2$-norm as $\theta^{(j)}$
			\Statex  $$\delta^{(j)} \gets \frac{\delta^{(j)}}{||\delta^{(j)}||}  ||\theta^{(j)}|| \;,\;\eta^{(j)} \gets \frac{\eta^{(j)}}{||\eta^{(j)}||}  ||\theta^{(j)}|| $$				
		\EndFor
		\Statex
		\Statex {// generate loss value at each $\alpha$ and $\beta$ location}
		\For{$\alpha$ in $[\alpha_{min},\ldots, \alpha_{max}]$}
			\For{$\beta$ in $[\beta_{min}, \ldots, \beta_{max}]$}
				\For{each layer $j$ of the model's layers}
					\State $\hat{\theta}^{(j)} \gets \theta^{(j)} + \alpha\delta^{(j)} + \beta\eta^{(j)}$
				\EndFor
				
				\State \texttt{loss} $\gets 0$
				\For{each sample $k$ in $D^{bon} \cup D^{mal} \cup D^{adv}$}
					\State \texttt{loss} $+= L(\hat{\theta}, \vx^{(k)}, y^{(k)})$
				\EndFor
				\State plot \texttt{avg(loss)} at coordinate ($\alpha$, $\beta$)
			\EndFor
		\EndFor
 	\end{algorithmic}
\end{algorithm}

\paragraph{Results.} As mentioned earlier, Figure~\ref{landscape_one_evasion} shows the landscape of the loss function based on a subset of the corresponding training sets (denoted by model-dataset in the figure) of the five models, while Figure~\ref{landscape_all_evasion} shows the same based on the union of these subsets (denoted by union-dataset in the figure): as it was trained naturally, the loss landscape of the \texttt{Natural} model (top subplot of Figure~\ref{landscape_one_evasion}) is associated with the standard generalization~\citep{filterwise}, and therefore one can not comment on the association of its flatness to robust generalization. When the adversarial variants from all the inner maximizers are incorporated in the \texttt{Natural}'s loss landscape (top subplot of Figure~\ref{landscape_all_evasion}), its chaotic structure clearly supports that standard and robust generalization are two different notions~\citep{madry2018generalization}.

On the other hand, we observe that the landscape's flatness and smoothness of the top most robust hardened models (\rmfgsm, \mBGA, and \dmfgsm) persist through their corresponding subplots in both Figures~\ref{landscape_one_evasion} and~\ref{landscape_all_evasion}. Loss landscape of the poorly hardened model with \mBCA~(bottom subplot of Figure~\ref{landscape_one_evasion}) shows a small bump near model's parameters and it gets more chaotic when the rest of the adversarial variants are incorporated (bottom subplot of Figure~\ref{landscape_all_evasion}). 

Note that without the knowledge of other inner maximizers, one can only generate a subplot similar to those of Figure~\ref{landscape_one_evasion} based on the considered inner maximizer, and we saw from these subplots that a bumpy geometry of loss landscape could be an indicator of a poorly hardened model (\mBCA). However, they still do not help in ranking the rest of the hardened models: \rmfgsm, \mBGA, and \dmfgsm. With the hope that it will convey a clearer picture, we are motivated to visualize the decision map of these models, as discussed next.

\subsection{Decision Map}

\begin{figure*}[!htb]
	\centering
	\resizebox{\textwidth}{!}{
	\begin{tabular}{|c|cccc|}
		\toprule
		 &  Loss Landscape (Model-Dataset) & Decision boundary (Model-Dataset) &  Loss Landscape (Union-Dataset) & Decision boundary (Union-Dataset)\\
		\toprule
		\raisebox{-12cm}{\rotatebox[origin=l]{90}{\mBCA \hspace{2.5cm} \mBGA \hspace{2.15cm}\rmfgsm \hspace{2.15cm}\dmfgsm \hspace{2.1cm}\texttt{Natural}}}  & 
	\begin{subfigure}[t]{1.7in}
		\includegraphics[width=1.7in]{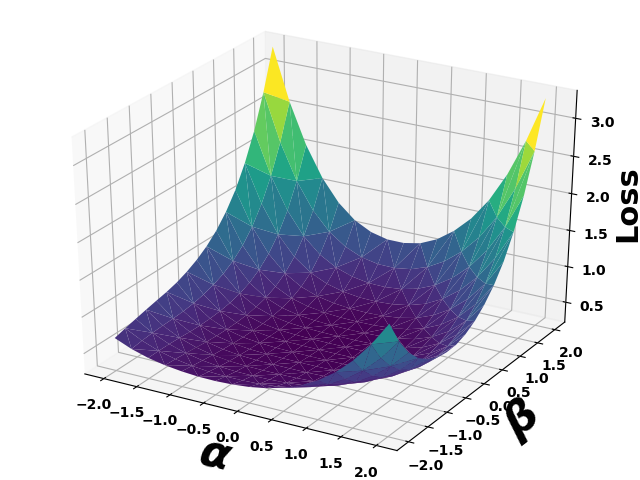}
		\includegraphics[width=1.7in]{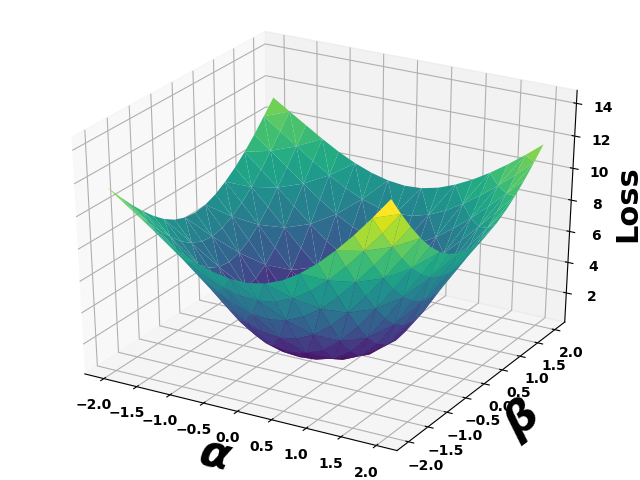}
		\includegraphics[width=1.7in]{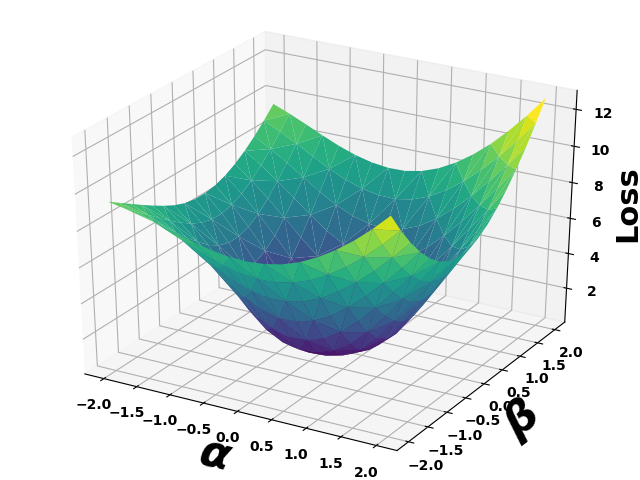}
		\includegraphics[width=1.7in]{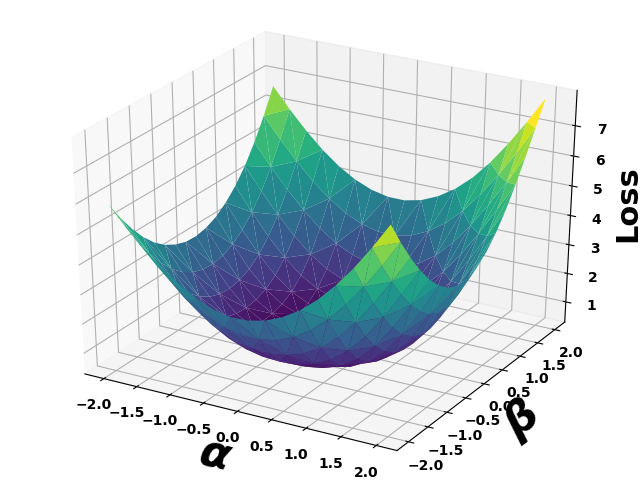}
		\includegraphics[width=1.7in]{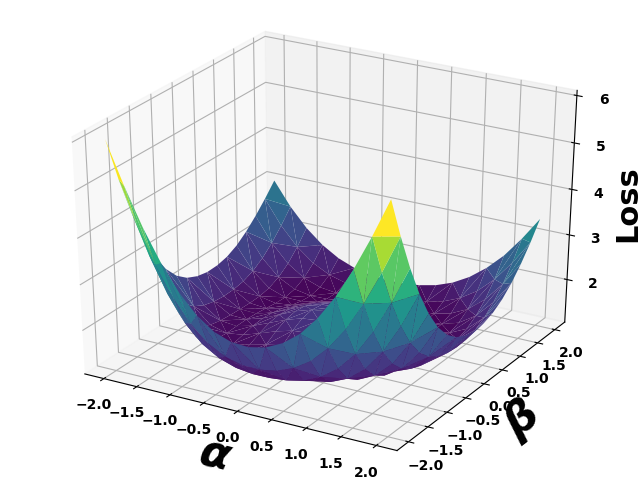}
		\caption{}
		\label{landscape_one_evasion}
	\end{subfigure} &
	\begin{subfigure}[t]{1.7in}
		\includegraphics[width=1.7in]{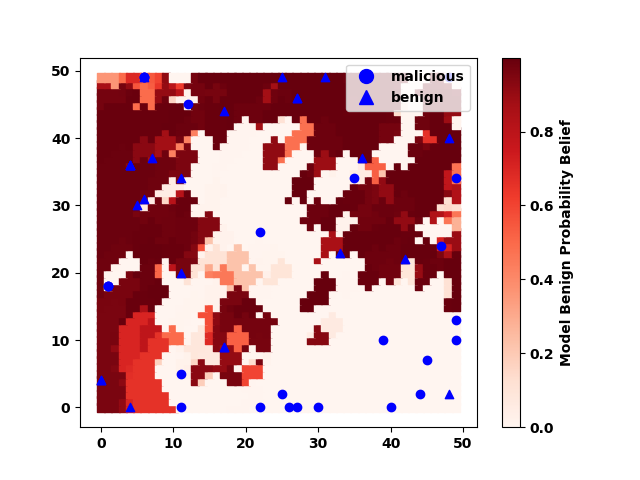}
		\includegraphics[width=1.7in]{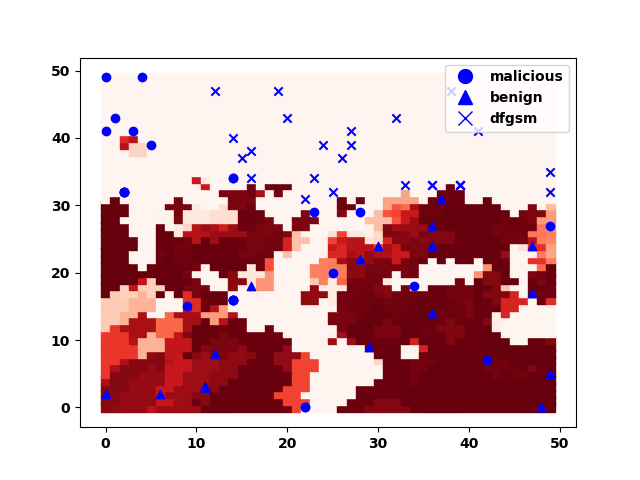}
		\includegraphics[width=1.7in]{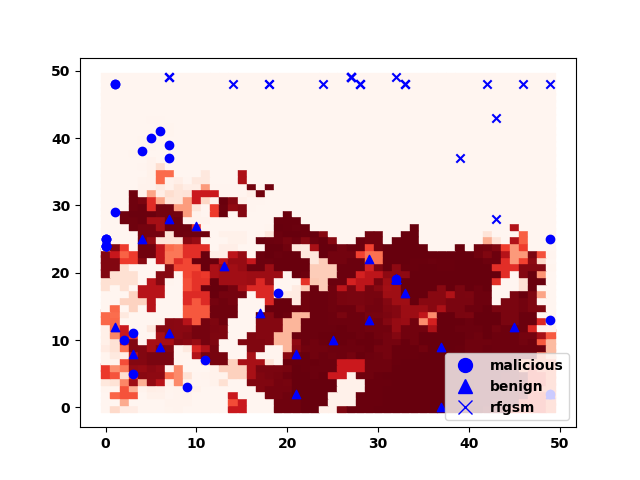}
		\includegraphics[width=1.7in]{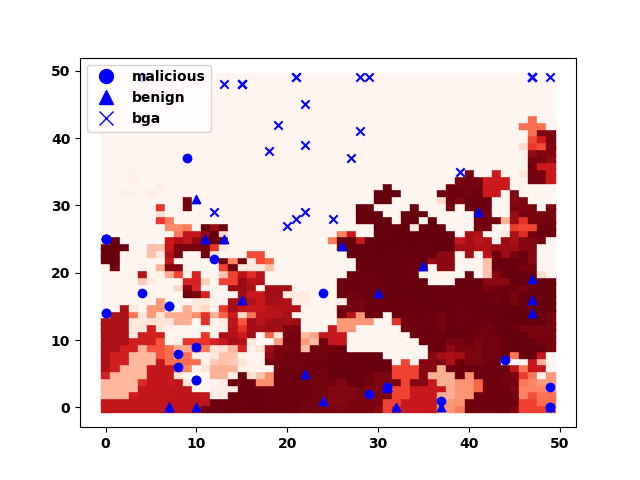}
		\includegraphics[width=1.7in]{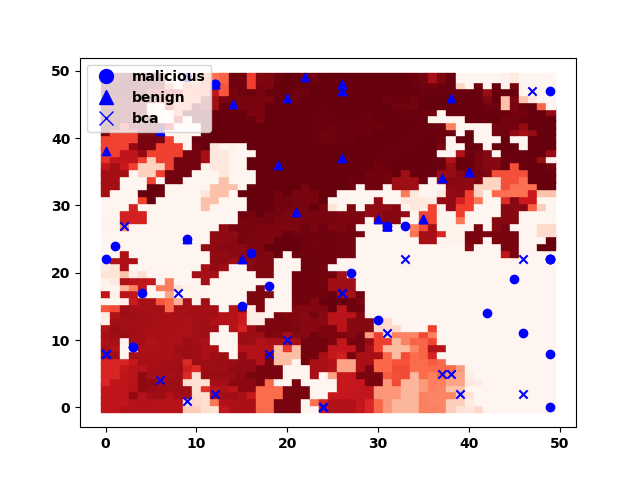}
		\caption{}
		\label{som_one_evasion}
	\end{subfigure} &
	\begin{subfigure}[t]{1.7in}
		\includegraphics[width=1.7in]{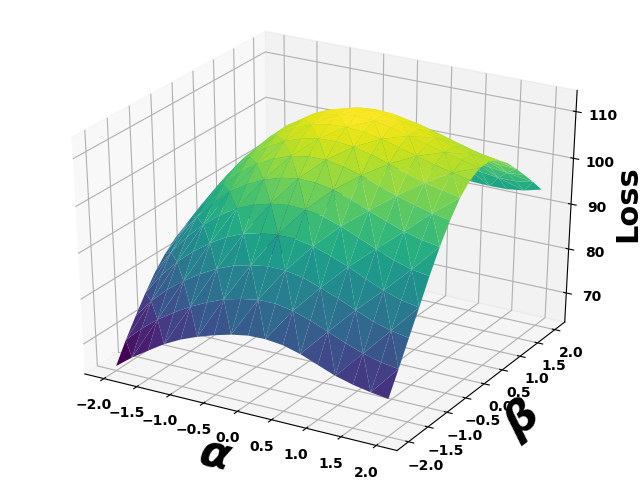}
		\includegraphics[width=1.7in]{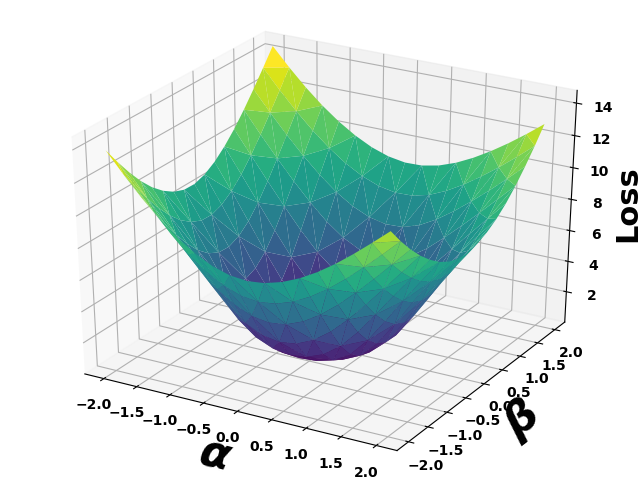}
		\includegraphics[width=1.7in]{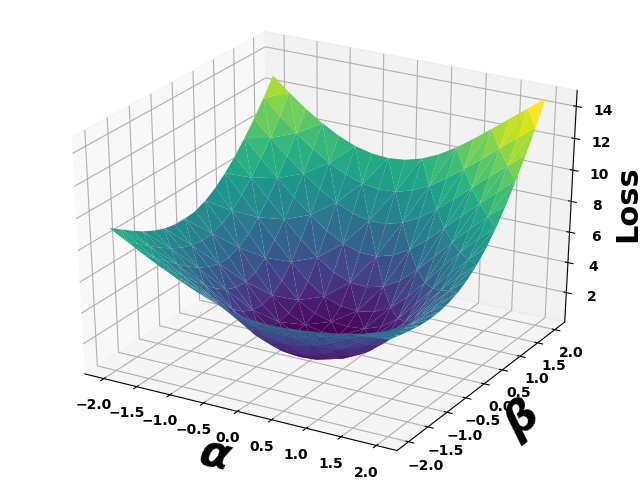}
		\includegraphics[width=1.7in]{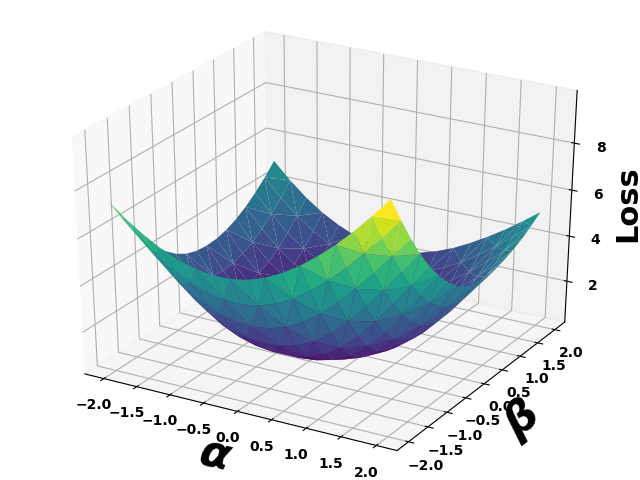}
		\includegraphics[width=1.7in]{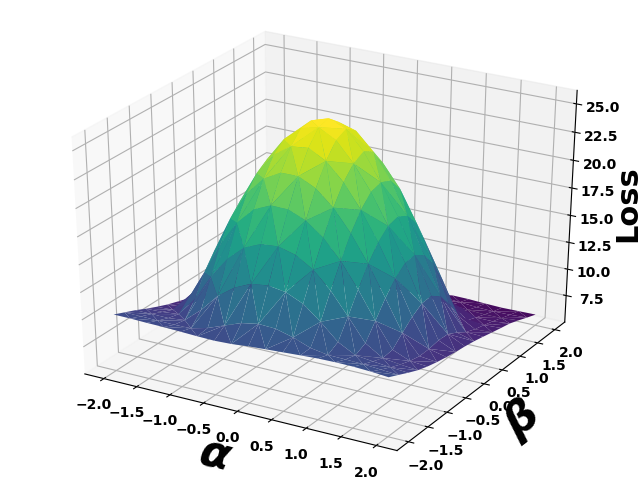}
		\caption{}
		\label{landscape_all_evasion}
	\end{subfigure} &
	\begin{subfigure}[t]{1.7in}
		\includegraphics[width=1.7in]{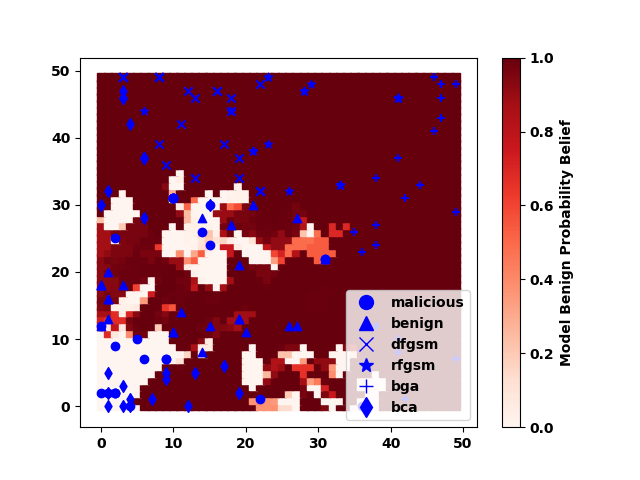}
		\includegraphics[width=1.7in]{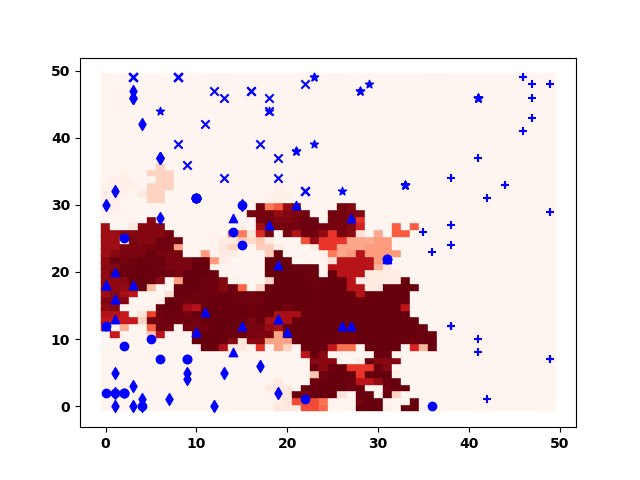}
		\includegraphics[width=1.7in]{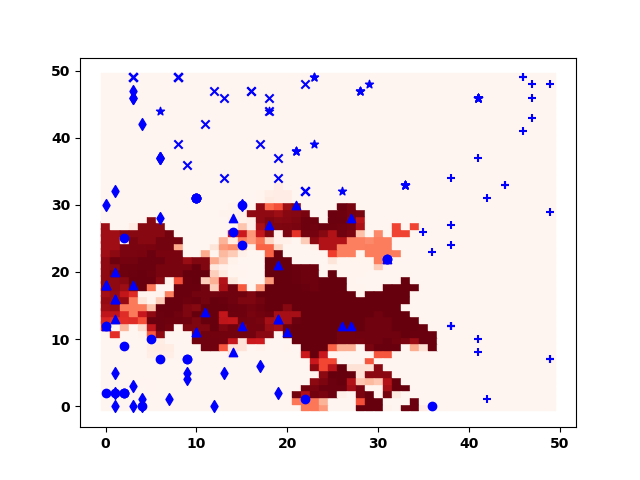}
		\includegraphics[width=1.7in]{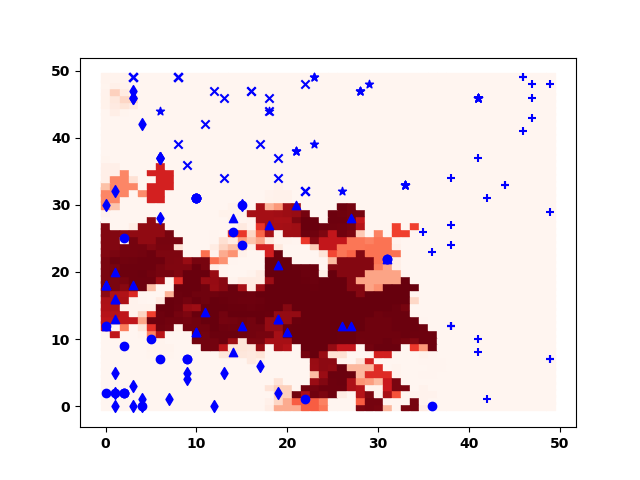}
		\includegraphics[width=1.7in]{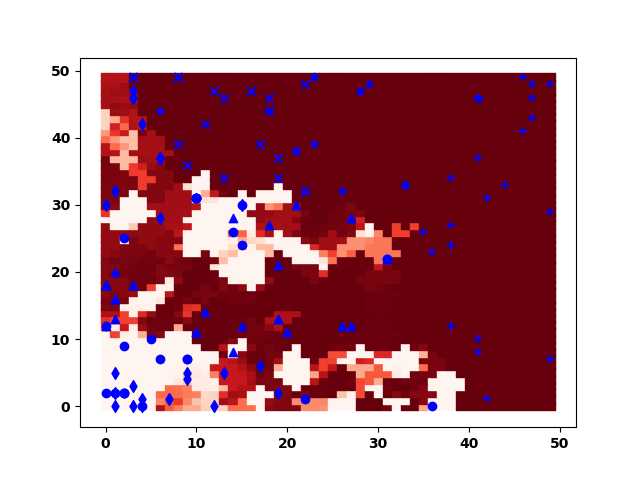}
		\caption{}
		\label{som_all_evasion}
	\end{subfigure} \\
	\bottomrule
	\end{tabular}
}
	\caption{(a) Loss landscapes for each model using only adversarial samples generated with the same inner maximizer used in training, with no adversarial variants for the naturally trained model. (b) Decision boundaries with a custom trained SOM per model using only malicious ($\Circle$), benign ($\triangle$), and vectors generated with the same adversarial ($\times$) inner maximizer used to train the model. The color indicates the SOM model's benign probability belief (white=low, red=high). (c) Loss landscapes for each model trained with malicious/benign and adversarial samples of each type. (d) Decision boundaries resulting from each type of adversarial training with all types of adversarial vectors: \dmfgsm~($\times$), \rmfgsm~($\star$), \mBGA~($+$), \mBCA~($\diamond$). The self-organizing map used is the same across all plots. \\ \\
	Flatness in the loss landscape when only considering adversarial variants generated using the training method (Column \ref{landscape_all_evasion}) does not imply model robustness, e.g.~the \mBCA trained model in in Column \ref{landscape_all_evasion} is flat but not robust. In Column \ref{landscape_all_evasion}, where all adversarial methods are considered, flatness does correlate with robustness which is also reinforced in Column \ref{som_all_evasion}.}
	\label{loss_landscape_and_som}
\end{figure*}

We use Self-Organizing Maps (\texttt{SOMs})  to visualize the decision map of natural and hardened models and superimpose samples and variants on it. This make false positives and negatives easy to see. A self-organizing map is a neural network trained using unsupervised learning to map a set of inputs to a lower dimensional mapping \citep{som_paper}. For input mapping we select either a model's training samples plus its adversarial variants (column 2 of Figure~\ref{som_one_evasion}) or training samples as well as the union of variants from each adversarial generation method (column 4 of Figure~\ref{som_one_evasion}). 

\paragraph{Method.} We use the {Somoclu} package \citep{somoclu} for training each \texttt{SOM}.  For all plots, the self-organizing map is a 50 by 50 grid of neurons. We train the map for 25 epochs on a dataset composed of 1,000 samples of each sample and variant type. After training on natural and adversarial samples for sufficient epochs, we  use the weight vectors of the neurons to plot the model's decision map in the lower dimensional mapping by feeding each to the model and color intensity coding the network's probabilistic belief in the input being \textit{benign}. We  next pass the samples and variants through the mapping and superimpose them on top of the implied decision map. 

\begin{algorithm}
	\tiny
	\caption{\tiny \\ \hfill Decision Boundary with Self-Organizing Maps \\
		\textbf{Requires:} \\
		$~$\hspace{\algorithmicindent}$\theta\in \mathbb{R}^{p}$: model parameters with $\theta^{(j)}\in \mathbb{R}^{p_j}$ as \\
		$~$\hspace{\algorithmicindent}$~$\hspace{\algorithmicindent} the  $j^{th}$ layer's parameters\\
		$~$\hspace{\algorithmicindent}$D^{bon}$: benign PEs dataset \\
		$~$\hspace{\algorithmicindent}$D^{mal}$: malicious PEs dataset \\
		$~$\hspace{\algorithmicindent}$D^{adv}$: adversarial malicious PEs dataset\\
       $~$\hspace{\algorithmicindent}$m$: feature vector size\\
}
	\label{som_code}
	\begin{algorithmic}[1]
	    \Statex // instantiate and train self-organizing map \texttt{SOM}
		\State \texttt{SOM} $\gets$ a 2D grid of neurons with weight vectors $\mathbf{w}\in \mathbb{R}^m$
		\For{each sample $i$ in  $D^{bon} \cup D^{mal} \cup D^{adv}$}
			\State train \texttt{SOM} with $\vx^{(i)}$
		\EndFor
		\Statex
		\Statex // plot the model's decision map
		\For{each neuron $j$ in \texttt{SOM}}
			\Statex $~$\hspace{\algorithmicindent}// compute benign probability at the neuron's vector
			\State calculate $p(y=0|\vx=\mathbf{w}^{(j)}, \theta)$
			\State plot the probability value at $j$'s coordinates in \texttt{SOM}
		\EndFor
		\For{each sample $i$ in  $D^{bon} \cup D^{mal} \cup D^{adv}$}
		\Statex $~$\hspace{\algorithmicindent}// get the best matching neuron
		\State $j \gets \arg\min_{j^\prime \in \texttt{SOM}} ||\vx^{(i)} - \mathbf{w}^{(j^\prime)} ||$ 
		\State Mark $i$ at $j's$ coordinates in \texttt{SOM}
		\EndFor 	
	\end{algorithmic}
\end{algorithm}

\paragraph{Results.} In the decision boundaries shown in Figures~\ref{som_one_evasion} and \ref{som_all_evasion}, dark red indicates that the model has a high confidence in the sample being benign. While the decision map based on model-dataset  of the \texttt{Natural} model (top subplot of Figure~\ref{som_one_evasion}) is fairly balanced between benign and malicious samples, the model's vulnerability to adversarial variations is clearly shown on its decision map based on union-dataset (top subplot of Figure~\ref{som_all_evasion}): the bulk of the adversarial variations are situated in regions that belong to the benign class with high confidence.

Interestingly, for \rmfgsm, the most robust model, the adversarial variations sit relatively far from the decision boundary between benign and malicious classes (middle subplot of Figure~\ref{som_one_evasion}), compared to the rest of the hardened models. As we move from more to less robust models, the adversarial variations get closer to the decision boundary. For instance, one can observe that close to the decision boundary: the \rmfgsm~model has one adversarial variation, the \mBGA~model has around three adversarial variations, and the \dmfgsm~model has around seven adversarial variations neighboring the decision boundary. It gets even worse for the poorly hardened model (\mBCA), where the adversarial variations step into regions that belong to the benign class with medium to high confidence (bottom subplot of Figure~\ref{som_one_evasion}). This observation is reaffirmed in Figure~\ref{som_all_evasion} where \mBCA's decision map is very similar to that of \texttt{Natural}. Furthermore, the regions with high-confidence benign class shrink as we move from less to more robust models and the adversarial variations from all the inner maximizers are situated in regions   that belong to the malicious class with high confidence.

\vspace{-.1in}

%% file: conclusions.tex
\section{Conclusion}
\label{sec:conclusion}

We provide a suite of visualization tools\footnote{The code will be made available upon publication} for insight into evaluating the effectiveness of adversarial hardening with the end goal of creating models resistant to adversarial methods. We investigated a variety of techniques that help explain how models benefit from adversarial (saddle-point) training. Using a dataset of PE files, we showed differences in loss progressions and loss histograms between naturally trained and adversarially trained models. We verified that supplementing model training with a single adversarial (inner maximizer) method provides resistance against the same method and sometimes other methods in the form of reduced loss values for adversarial variants. We also visualize the parameter space and input space of adversarially trained models using filter-wise normalization and self-organizing maps, respectively. We saw that the geometry  of the loss landscape of a hardened model may provide an insight about its robust generalization. Based on our experiments, it appears that decision boundary and its location relative to the adversarial variations has a stronger association with the hardened model's robustness, compared to the geometry of the loss landscape around the model's parameters. While this paper addressed models with binary feature space, in our future work, we would like to investigate the presented methods on models with continuous feature space (e.g., images).

\section*{Acknowledgments}

This work was supported by CrowdStrike, the MIT-IBM Watson AI Lab
and CSAIL CyberSecurity Initiative.